\pdfoutput=1

\documentclass[11pt]{article}

\usepackage[]{acl}

\usepackage{times}
\usepackage{latexsym}

\usepackage[T1]{fontenc}

\usepackage[utf8]{inputenc}

\usepackage{microtype}

\usepackage{amssymb} 
\usepackage{bbding} 
\newcommand{\RNum}[1]{\uppercase\expandafter{\romannumeral #1\relax}} 
\usepackage{graphicx} 
\usepackage{multirow} 
\usepackage{color} 

\usepackage{hhline}
\usepackage{boldline}
\usepackage{subfig}

\newcommand\eat[1]{}
\usepackage{amsmath} 
\usepackage{array} 
\usepackage{enumerate} 
\usepackage{booktabs} 
\usepackage{framed} 
\usepackage{color} 
\usepackage{savesym} 
\savesymbol{Cross} 
\usepackage{hyperref} 
\usepackage{marvosym} 

\title{\textsc{Wiki}Diverse: A Multimodal Entity Linking Dataset with Diversified Contextual Topics and Entity Types}

\author{Xuwu Wang\textsuperscript{1}, Junfeng Tian\textsuperscript{2}, Min Gui\textsuperscript{3}\thanks{\;\;This work was conducted when Min Gui worked at Alibaba.}, Zhixu Li\textsuperscript{1\textrm{\Letter}}, Rui Wang\textsuperscript{4},\\ 
\bf Ming Yan\textsuperscript{2}, \bf Lihan Chen\textsuperscript{1}, \bf Yanghua Xiao\textsuperscript{1,5\textrm{\Letter}}\\
\textsuperscript{1} School of Computer Science, Fudan University, China \\
\textsuperscript{2} Alibaba Group, China \
\textsuperscript{3} Shopee, Singapore \
\textsuperscript{4} Vipshop (China) Co., Ltd., China \\
\textsuperscript{5} Fudan-Aishu Cognitive Intelligence Joint Research Center, China\\
\tt \{xwwang18,zhixuli,shawyh\}@fudan.edu.cn,\\
\tt \{tjf141457, ym119608\}@alibaba-inc.com, \\
\tt min.gui@shopee.com, mars198356@hotmail.com, lhc825@gmail.com}

\begin{document}
\maketitle
\begin{abstract}
Multimodal Entity Linking (MEL) which aims at linking mentions with multimodal contexts to the referent entities from a knowledge base (e.g., Wikipedia), is an essential task for many multimodal applications. Although much attention has been paid to MEL, the shortcomings of existing MEL datasets including limited contextual topics and entity types, simplified mention ambiguity, and restricted availability, have caused great obstacles to the research and application of MEL. In this paper, we present \textsc{Wiki}Diverse, a high-quality human-annotated MEL dataset with diversified contextual topics and entity types from Wikinews, which uses Wikipedia as the corresponding knowledge base. A well-tailored annotation procedure is adopted to ensure the quality of the dataset. Based on \textsc{Wiki}Diverse, a sequence of well-designed MEL models with intra-modality and inter-modality attentions are implemented, which utilize the visual information of images more adequately than existing MEL models do. Extensive experimental analyses are conducted to investigate the contributions of different modalities in terms of MEL, facilitating the future research on this task. The dataset and baseline models are available at {https://github.com/wangxw5/wikiDiverse}.
\end{abstract}

\section{Introduction}\label{sec:intro}
\textbf{Entity linking (EL)} has attracted increasing attention in the natural language processing community, which aims at linking ambiguous mentions to the referent unambiguous entities in a given knowledge base (KB) \cite{shen2014entity}. It has been applied to a lot of downstream tasks such as information extraction \cite{yaghoobzadeh2016noise}, question answering \cite{yih2015semantic} and semantic search \cite{blanco2015fast}. 

\begin{figure}[t]
	\centering
	\includegraphics[width=1.0 \columnwidth]{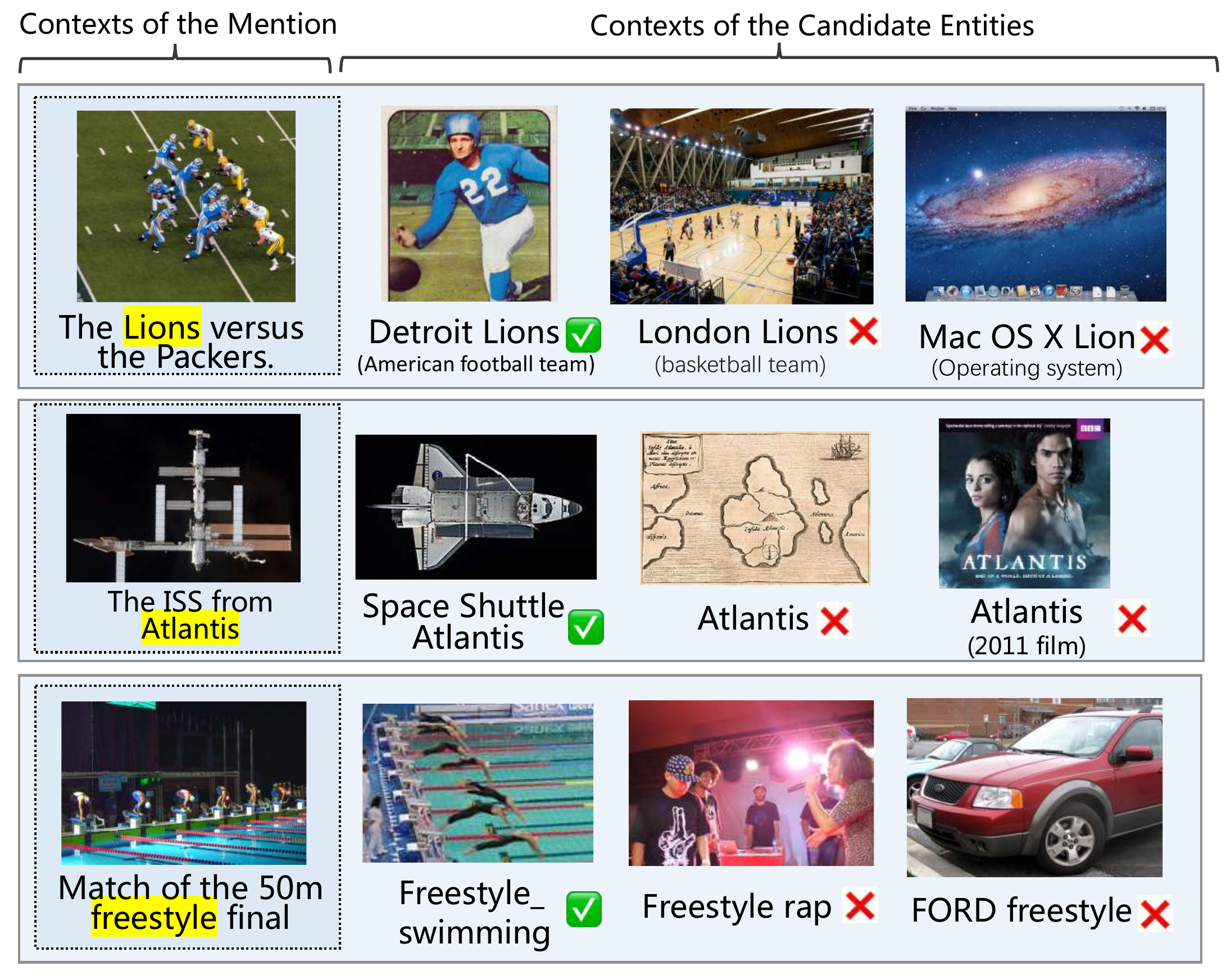}
	\caption{Several MEL examples with mentions highlighted in the caption and the first entity of each entity listed as the gold label.}
	\label{fig:example}
\end{figure}

\begin{table*}[htbp]
\scriptsize
\setlength\tabcolsep{0.07cm}
\centering
\begin{tabular}{l|l|ccccccccc}
\toprule
\textbf{Task} &  \textbf{Dataset} & \textbf{Source} & \textbf{KB} & \textbf{Modality} & \textbf{Topic} & \textbf{Ent. Types} & \textbf{Manual} & \textbf{Open} & \textbf{Lang} & \bf Size \\ \midrule

 & AIDA\cite{hoffart-etal-2011-robust} & News & Wikipedia & $T_m\rightarrow T_e$ & Multiple & Multiple & \CheckmarkBold & \CheckmarkBold & en & 1K docs\\
 & MSNBC\cite{cucerzan-2007-large} & News & Wikipedia & $T_m\rightarrow T_e$ & Multiple & Multiple & \CheckmarkBold & \CheckmarkBold & en & 20 docs\\
 & AQUA\cite{milne2008learning} & News & Wikipedia & $T_m\rightarrow T_e$ & Multiple & Multiple & \CheckmarkBold & \CheckmarkBold & en & 50 docs\\
 & ACE2004\cite{ratinov-etal-2011-local} & News & Wikipedia & $T_m\rightarrow T_e$ & Multiple & Multiple & \CheckmarkBold & \CheckmarkBold & en & 57 docs\\
 & CWEB\cite{guo2018robust} & Web & Wikipedia & $T_m\rightarrow T_e$ & Multiple & Multiple & \XSolidBrush & \CheckmarkBold & en & 320 docs\\
 & WIKI\cite{guo2018robust} & Wiki & Wikipedia & $T_m\rightarrow T_e$ & Multiple & Multiple & \XSolidBrush & \CheckmarkBold & en & 320 docs\\
\multirow{-7}{*}{\textbf{EL}} & Zeshel\cite{logeswaran-etal-2019-zero} & Wiki & Wikia & $T_m\rightarrow T_e$ & Multiple & Multiple & \XSolidBrush & \CheckmarkBold & en & -\\ \hline
 & Snap\cite{moon-etal-2018-multimodal-named} & Social Media & Freebase & $T_m,V_m\rightarrow T_e$ & Multiple & Multiple & \CheckmarkBold & \XSolidBrush & en & 12K captions\\
 & Twitter\cite{adjali-etal-2020-building} & Social Media & Twitter users & $T_m,V_m\rightarrow T_e,V_e$ & Multiple & PER, ORG & \XSolidBrush & \XSolidBrush & en & 4M tweets\\
 & Movie\cite{gan2021multimodal} & Movie Reviews & Wikipedia & $T_m,V_m\rightarrow T_e,V_e$ & Movie & PER & \CheckmarkBold & \CheckmarkBold & en & 1K reviews\\
 & Weibo\cite{zhang2021attention} & Social Media & Baidu Baike & $T_m,V_m\rightarrow T_e,V_e$ & multiple & PER & \XSolidBrush & \CheckmarkBold & cn & 25K posts\\ \cline{2-11}
\multirow{-5}{*}{\textbf{MEL}} & \textsc{Wiki}Diverse & News & Wikipedia & $T_m,V_m\rightarrow T_e,V_e$ & Multiple & Multiple & \CheckmarkBold & \CheckmarkBold & en & 8K captions\\ \bottomrule
\end{tabular}
\caption{Overview of EL and MEL datasets. $T_m$ ($T_e$) and $V_m$ ($V_e$) represent the textual and visual contexts of mentions $m$ (or entities $e$) respectively, ``Manual'' denotes whether it is manually annotated, and ``Open'' denotes whether it is an open source.}
\label{tbl:existing_datasets}
\end{table*}

\begin{figure*}[t]
	\centering
	\includegraphics[width=1.8 \columnwidth]{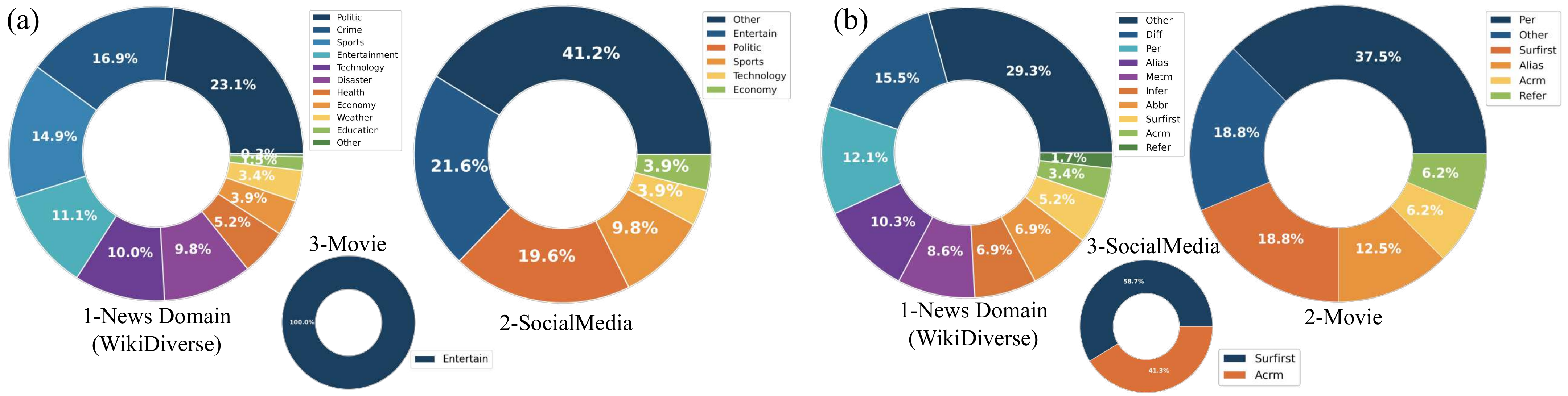}
	\caption{(a) compares the topic distribution of different domains. The statistics of social media are observed on sampled Twitter~\cite{adjali-etal-2020-building}. The statistics of news domain are observed on \textsc{Wiki}Diverse. The statistics of Movie domain are observed on movie reviews sampled from IMDb. (b) compares the ambiguity distribution of different domains, where ten types of ambiguity are observed on our dataset, including different types of objects with the same name (Diff), persons with the same name (Per), Alias, metonymy (Metm), inferring (Infer), abbreviation (Abbr), surname or first name (SurFirst), acronym (Acrm), reference (Refer) and others.}
	\label{fig:topic}
\end{figure*}

As named entities (i.e., mentions) with multimodal contexts such as texts and images are ubiquitous in daily life, recent studies~\cite{moon-etal-2018-multimodal-named,adjali-etal-2020-building} turn their focus towards improving the performance of EL models through utilizing visual information, i.e., \textbf{Multimodal Entity linking (MEL)}\footnote{In this paper, we focus on mentions coming from text spans and leave the visual mentions (i.e. objects from the images) for the future work.}. 
Several MEL examples are depicted in Figure~\ref{fig:example}, where the images could effectively help the disambiguation for entity mentions of different types.
Due to its importance to many multimodal understanding tasks including VQA, multimodal retrieval, and the construction of multimodal KBs, much effort has been dedicated to the research of MEL.
\citet{moon-etal-2018-multimodal-named} first addressed the MEL task under the zero-shot setting. \citet{adjali-etal-2020-building} designed a model to combine the visual, textual and statistical information for MEL. \citet{zhang2021attention} designed a two-stage mechanism that first determines the relations between images and texts to remove negative impacts of noisy images and then performs the disambiguation. \citet{gan2021multimodal} disambiguated visual mentions and textual mentions respectively at first, and then used graph matching to explore possible relations among inter-modal mentions.

Although much attention has been paid to MEL, the existing MEL datasets as listed in the middle rows of Table~\ref{tbl:existing_datasets} have deficiencies in the following aspects, which hinder the further advancement of research and application for MEL.

\begin{itemize}
\item {\bf Limited Contextual Topics.} 
	As shown in Figure~\ref{fig:topic}(a), the existing MEL datasets are mainly collected from social media or movie reviews, where there are only 5 topics in the social media domain and 1 topic in the movie review domain. But as we observed in the news domain, there are more than 10 topics including other popular topics like disaster and education. The lack of topics would limit the generalization ability of the MEL model.
	
	\item {\bf Limited Entity Types.} Entities in the existing MEL datasets mainly belong to the types of ``person (PER)'' and ``organization (ORG)''. This restricts the application of the MEL models over other entity types such as locations, events, etc., which are also ubiquitous in common application scenarios.

	\item {\bf Simplified Mention Ambiguity}: Some datasets such as Twitter~\cite{adjali-etal-2020-building} create artificial ambiguous mentions by replacing the original entity names with the surnames of persons or acronyms of organizations. Besides, limited entity types also lead to the limited mention ambiguity that only occurs with PER and ORG.
	According to our statistics of different domains as depicted in Figure~\ref{fig:topic}(b), there are overall ten kinds of mention ambiguities in news domain such as Wikinews\footnote{https://www.wikinews.org. It is a free-content news wiki.}, while existing datasets collected from social media or movie reviews only cover a small scope of ambiguity.

	\item {\bf Restricted Availability.} Most of the existing MEL datasets are not publicly available.
\end{itemize}

To enable more detailed research of MEL, we propose a manually-annotated MEL dataset named \textsc{Wiki}Diverse with multiple topics and multiple entity types. It consists of 8K image-caption pairs collected from WikiNews and is based on the KB of Wikipedia with \textasciitilde16M entities in total. Both the mentions and entities are characterized by multimodal contexts. We design a well-tailored annotation procedure to ensure the quality of \textsc{Wiki}Diverse and analyze the dataset from multiple perspectives (Section~\ref{sec:dataset}). 
Based on \textsc{Wiki}Diverse, we propose a sequence of MEL models with intra-modality and inter-modality attentions, which utilize the visual information of images more adequately than the existing MEL models (Section~\ref{sec:experiment}). 
Furthermore, extensive empirical experiments are conducted to analyze the contributions of different modalities for the MEL task and visual clues provided by the visual contexts (Section~\ref{sec:analysis}).
In summary, the contributions of our work are as follows:
\begin{itemize}
  \item We present a new manually annotated high-quality MEL dataset that covers diversified topics and entity types.
  \item Multiple well-designed MEL models with intra-modal attention and inter-modal attention are given which could utilize the visual information of images more adequately than the previous MEL models.
  \item Extensive empirical results quantitatively show the role of textual and visual modalities for MEL, and detailed analyses point out promising directions for the future research.

\end{itemize}

\section{Related Work}\label{sec:related_work}
\paragraph{Textual EL} 
There is vast prior research on textual entity linking. 
Multiple datasets have been proposed over the years including the manually-annotated high-quality datasets like AIDA \cite{hoffart-etal-2011-robust}, automatically-annotated large-scale datasets like CWEB \cite{guo2018robust} and zero-shot datasets like Zeshel \cite{logeswaran-etal-2019-zero}. To evaluate the EL models' performance, it is usual to train on the AIDA-train dataset, and test on the datasets of AIDA-test, MSNBC\cite{cucerzan-2007-large}, AQUAINT\cite{milne2008learning}, etc. However, as mentioned in \cite{de2020autoregressive}, many methods have achieved high and similar results within recent three years. One possible explanation is that it may simply be near the ceiling of what can be achieved for these datasets, and it is difficult to conduct further research based on them.

\paragraph{Multimodal EL} 
In recent years, the growing trend towards multimodality requires to extend the research of EL from monomodality to multimodality.
\citet{moon-etal-2018-multimodal-named} first address the MEL task and build a zero-shot framework, which extracts textual, visual and lexical information for EL in social media posts. However, its proposed dataset is unavailable due to GDPR rules.
\citet{adjali-etal-2020-building,10.1007/978-3-030-45439-5_31} propose a framework of automatically building the MEL dataset from Twitter. The dataset has limited entity types and ambiguity of mentions, thus it is not challenging enough. 
\citet{zhang2021attention} study on a Chinese MEL dataset collected from the Chinese social media platform Weibo, which mainly focuses on the person entities.
\citet{gan2021multimodal} release a MEL dataset collected from movie reviews and propose to disambiguate both visual and textual mentions. This dataset mainly focuses on characters and persons of the movie domain.
\citet{wang2021ccks} propose three MEL datasets, which are built from Weibo, Wikipedia, and Richpedia information and use CNDBpedia, Wikidata and Richpedia as the corresponding KBs. However, using Wikipedia as the target dataset may lead to the data leakage problem as many language models are pretrained on it. 

Our MEL dataset is also related to other named entity-related multimodal datasets, including entity-aware image caption datasets \cite{biten2019good,tran2020transform,liu-etal-2021-visual}, multimodal NER datasets \cite{zhang2018AAAI,lu-etal-2018-visual}, etc.
However, the entities in these datasets are not linked to a unified KB.
So our research of MEL can enhance the understanding of named entities, thereby enhancing the research in these areas.

\section{Problem Formulation}\label{sec:formulation}
Multimodal entity linking is defined as mapping a mention with multimodal contexts to its referent entity in a pre-defined multimodal KB. 
Since the boundary and granularity of mentions may be controversial, the mention span is usually pre-specified.
Here we assume each mention has a corresponding entity in the KB, which is the \emph{in-KB} evaluation problem.

Formally, let $E$ represent the entity set of the KB, which usually contains millions of entities. Each mention $m$ or entity $e_i\in E$ is characterized by the corresponding visual context $V_m, V_{e_i}$ and textual context $T_m, T_{e_i}$. Here $T_m$ and $T_{e_i}$ represent the textual spans around $m$ and $e_i$ respectively. $V_m$ is the image correlated with $m$ and $V_{e_i}$ is the image of $e_i$ in the KB. In real life, entities in KBs may contain more than one image. To simplify it, we select the first image of $e_i$ as $V_{e_i}$ and leave MEL with multiple images per entity as the future work. So the referent entity of mention $m$ is predicted through:
$$
    e^*(m)=\mathop{\arg\max}_{e_i\in E} \ \ \Psi\left(m\left(T_m,V_m\right);e_i\left(T_{e_i},V_{e_i}\right)\right).
$$
where $\Psi(\cdot)$ represents the similarity score between the mention and entity.

\section{Dataset Construction}\label{sec:dataset}
In this section, we present the dataset construction procedure. Many factors including annotation quality, coverage of topics, diversity of entity types, coverage of ambiguity are taken into consideration to ensure the research value of \textsc{Wiki}Diverse.

\subsection{Data Collection}

\paragraph{Data Source Selection}  
1) For the source of image-text pairs, considering news articles are widely-studied in traditional EL \cite{hoffart-etal-2011-robust, cucerzan-2007-large} and usually cover a wide range of topics and entity types, we decide to use news articles. Wikinews and BBC are two popular sources of news articles. So we compared them from two aspects. As shown in Table~\ref{tab:comparison}, Wikinews has advantages in terms of alignment degree between image-text pairs and MEL difficulty. So we select the image-caption pairs of Wikinews to build the corpus. 
2) For the source of KB, we use the commonly-used Wikipedia \cite{hoffart-etal-2011-robust,ratinov-etal-2011-local,guo2018robust}. We also provide the annotation of the corresponding Wikidata entity for flexible studies.

\begin{table}[t]
\small
\setlength{\tabcolsep}{1.5pt}
\centering
\begin{tabular}{l|ccc|ccc}
\toprule
\multirow{2}{*}{Source} & \multicolumn{3}{c|}{Alignment Degree with Image} & \multicolumn{3}{c}{MEL Difficulty} \\
\cline{2-7}
                        & Caption      & Headline     & First Sent.     & No        & Easy        & Hard       \\
\midrule
Wikinews                & 99\%          & 30\%          & 23\%       & 1\%          & 5\%         & 94\%       \\
BBC                     & 82\%          & 53\%          & 53\%       & 2\%          & 30\%        & 68\% \\
\bottomrule
\end{tabular}
\caption{Comparing the alignment degrees and corresponding MEL difficulty of image-caption, image-news headline, and image-first sentence between Wikinews and BBC, where  the MEL difficulty is measured through the surface form similarity between mentions and entities.}
\label{tab:comparison}
\end{table}

\paragraph{Data Acquisition} 1) For the image-caption pairs, we collect all the English news from the year 2007 to 2020 from Wikinews with multiple topics including {sports, politics, entertainment, disaster, technology, crime, economy, education, health and weather}. The data cover most of the common topics in the real world. Finally, we obtain a raw corpus with 14k image-caption pairs. 2) For the KB, we use the Wikipedia\footnote{The Wikipedia dump of January 01, 2021}. The entity set consists of all the entities in the main namespace with the size of \textasciitilde16M. 

\paragraph{Data Cleaning}
For the image-caption pairs, we remove the cases that 1) contain pornographic, profane, and violent content; 2) the text is shorter than 3 words.
Finally, we get a corpus with 8K image-caption pairs.

\subsection{Annotation}

\paragraph{Annotation Design}

The primary goal of \textsc{Wiki}Diverse is to link mentions with multimodal contexts to the corresponding Wikipedia entity. 
Therefore, given an image-text pair, annotators need to 1) detect mentions from the text (Mention Detection, MD) and 2) label each detected mention with the corresponding entity in the form of a Wikipedia URL (Entity Linking, EL). For mentions that do not have corresponding entities in Wikipedia, they are labeled with ``NIL''. Seven common entity types (i.e., Person, Organization, Location, Country, Event, Works, Misc) are required to be annotated. To avoid subjective errors, we design detailed annotation guidelines with multiple samples to avoid the controversy of mention boundary, mention granularity, entity URL, etc. Details can be found in the Appendix. We also hold regular communications to discuss some emerging annotations problems.

\paragraph{Annotation Procedure}

The annotators include 13 annotators and 2 experienced experts. All annotators have linguistic knowledge and are instructed with detailed annotation principles. Each image-caption pair is independently annotated by two annotators. Then an experienced expert goes over the controversial annotations, and makes the final decision. 
Following \citet{ding2021few}, we calculate the Cohen’s Kappa to measure the agreements between two annotators. The Kappa of MD and EL are 88.98\% and 83.75\% respectively, indicating a high degree of consistency.

\begin{figure}[t]
    \centering
    \includegraphics[width=\columnwidth]{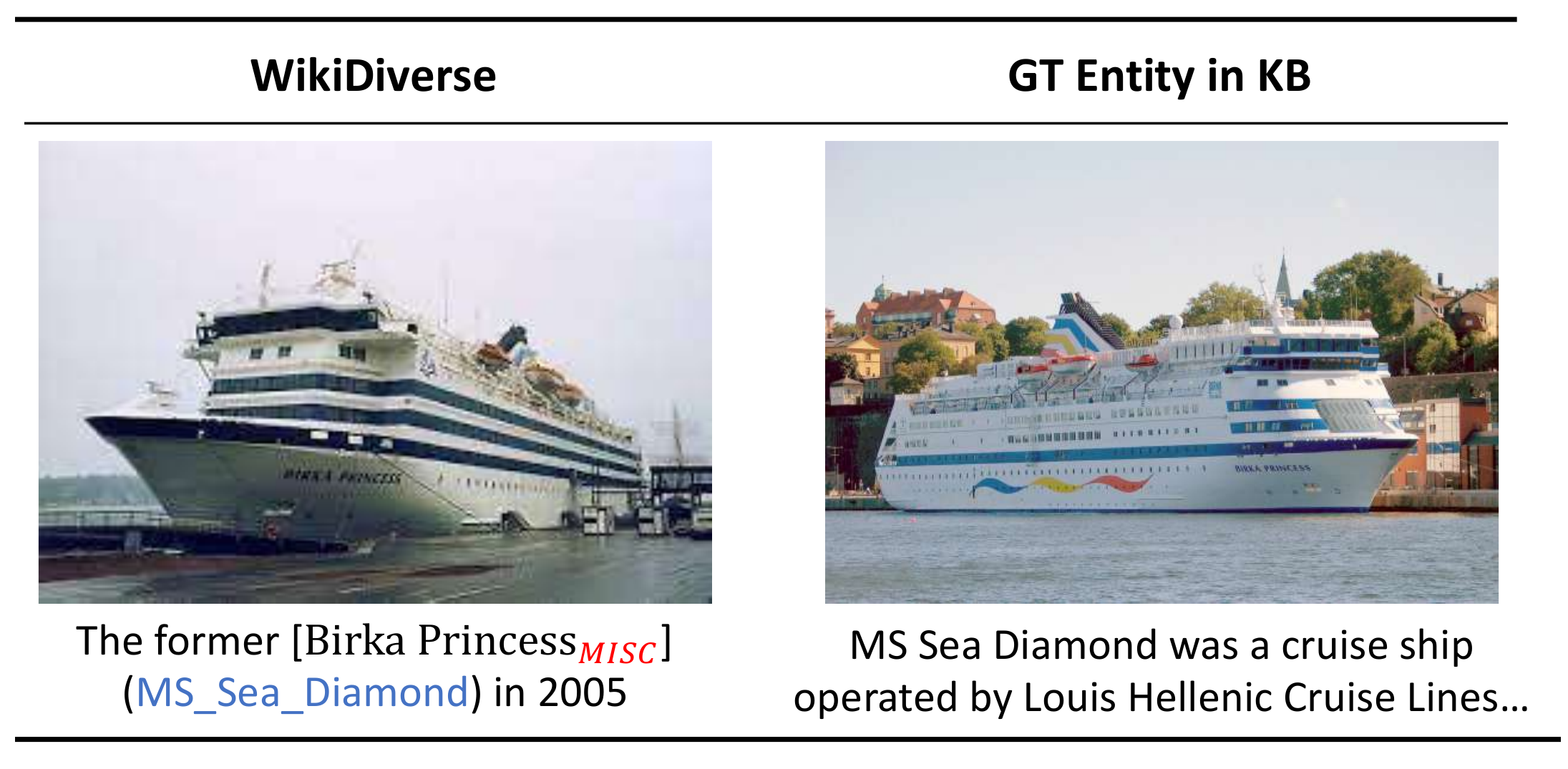}
    \caption{An example from \textsc{Wiki}Diverse. GT denotes the \textit{ground truth} entity. The red text and blue text indicate the annotated entity type and Wikipedia entity respectively.}
    \label{fig:eg_of_annotation}
\end{figure}

\subsection{Analysis of \textsc{Wiki}Diverse}

\begin{figure}[t]
	\centering
	\includegraphics[width=0.9\columnwidth]{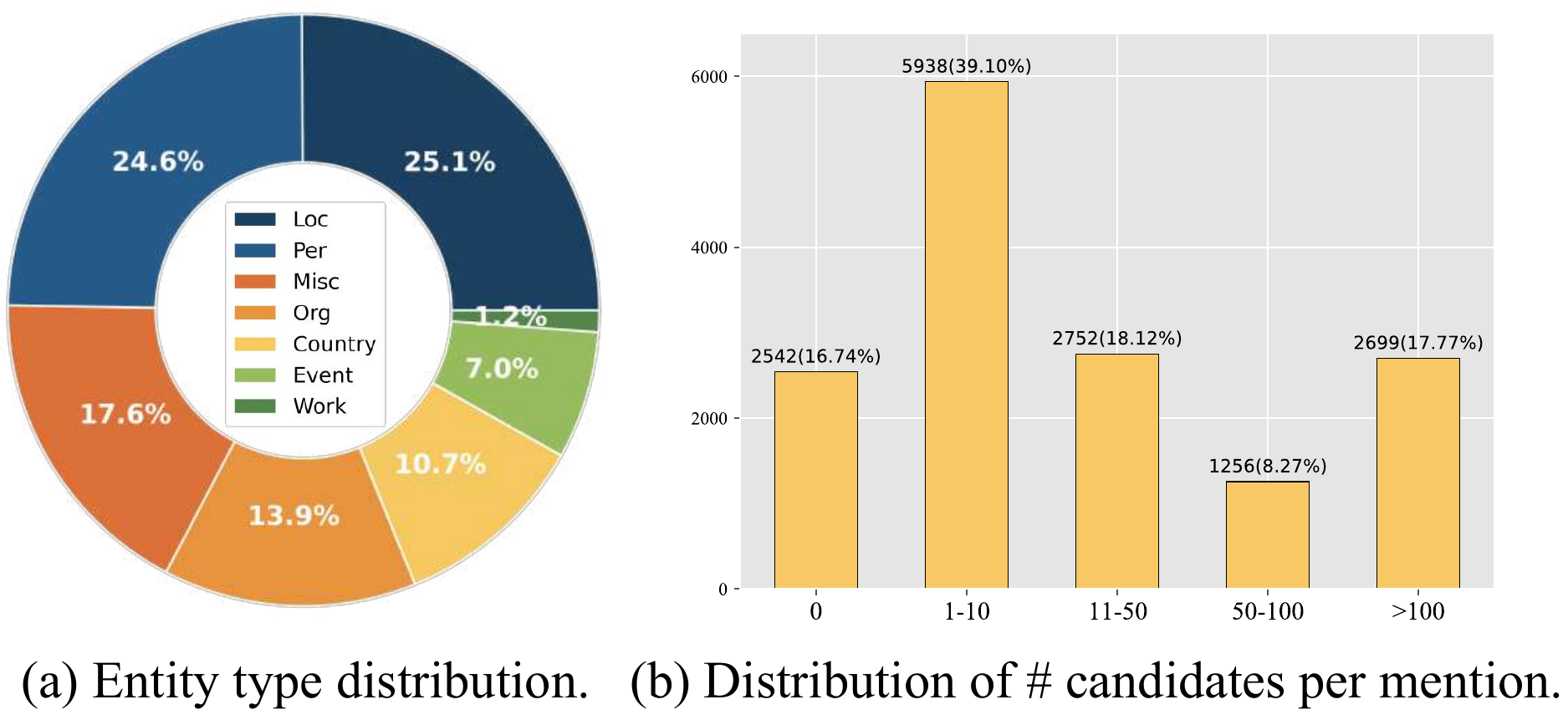}
	\caption{More statistics of \textsc{Wiki}Diverse. (a) Entity type distribution. (b) Distribution of the number of candidates per mention}
	\label{fig:distribution}
\end{figure}

\paragraph{Size and Distribution of \textsc{Wiki}Diverse}
We divide \textsc{Wiki}Diverse into training set, validation set, and test set with the ratio of 8:1:1. 
The statistics of \textsc{Wiki}Diverse are shown in Table~\ref{tbl:dataset_stats}. 
The collected Wikipedia KB  has \textasciitilde16M entities in total (i.e. $|E|\approx$16M).
Besides, we report the entity type distribution in Figure~\ref{fig:distribution}(a) and report the topic distribution in Figure~\ref{fig:topic}(a).
\begin{table}[t]
\setlength{\tabcolsep}{5pt}
\centering
\begin{tabular}{lcccc}
\toprule
 & \textbf{Train} & \textbf{Dev.} & \textbf{Test} & \textbf{Total} \\ 
\midrule
\# pairs & 6311 & 755 & 757 & 7823 \\
\# ment. per pair & 2.09 & 2.06 & 2.07 & 2.09 \\
\# words per pair & 10.16 & 10.30 & 10.03 & 10.16 \\ \bottomrule
\end{tabular}
\caption{Statistics of \textsc{Wiki}Diverse.}
\label{tbl:dataset_stats}
\end{table}

\paragraph{Difficulty Measure} 

Firstly, we compare surface form similarity of mentions and ground-truth entities. 51.31\% of the mentions have different surface forms compared with ground-truth entities. Specifically, 16.05\% of the mentions are totally different from the ground-truth entities. The large difference of the surface form brings challenges for MEL.

Secondly, we report the \#candidate entities for each mention in Figure~\ref{fig:distribution}(b). Intuitively, the more entities a mention may refer to, the more ambiguous the mention is, and the more difficult the EL/MEL is. Specifically, we generate a $m\rightarrow e$ hash list based on the ($m,e$) co-occurrence statistics from Wikipedia (See Section~\ref{sec:cd} for details). As shown in Figure~\ref{fig:distribution}(b), we can see that 1) 44.2\% mentions have more than 10 candidate entities. 
2) 16.7\% mentions are not contained in the hash list, which means their candidates are the entire entity set of the KB.

Thirdly, we randomly sample 200 image-caption pairs from \textsc{Wiki}Diverse to evaluate the diversity of ambiguity. As shown in Figure~\ref{fig:topic}(b), \textsc{Wiki}Diverse covers a wide range of ambiguity.

\section{Methods}\label{sec:experiment}

It is challenging to directly predict the entity from a large-scale KB because it consumes large amounts of time and space resources. Therefore, following previous work \cite{yamada-etal-2016-joint,ganea-hofmann-2017-deep,de2020autoregressive}, we split MEL into two steps: 1) candidate retrieval (CR) is first used to guarantee the recall and obtain a candidate entity set consisting of the TopK entities that are most similar to the mention; 2) entity disambiguation (ED) is then conducted to guarantee the precision and predict the entity with the highest matching score.

\subsection{Candidate Retrieval} \label{sec:cd}

Existing methods \cite{yamada-etal-2016-joint,ganea-hofmann-2017-deep,le-titov-2018-improving} mainly utilize two types of clues to generate the candidate entity set $E_m$: (\RNum{1}) the $m\rightarrow e$ hash list recording prior probabilities from mentions to entities: $P(e|m)$. (\RNum{2}) the similarity between the contexts of mention $m$ and entity $e$.

Following these works, we implement a series of baselines as follows: (\RNum{1}) $\mathbf{P(e|m)}$ \cite{ganea-hofmann-2017-deep}: $P(e|m)$ is calculated based on 1) mention entity hyperlink count statistics from Wikipedia; 2) Wikipedia redirect pages; 3) Wikipedia disambiguation pages. (\RNum{2}) \textbf{Baselines of textual modality}: we retrieve the TopK candidate entities with the most similar textual context of the mention based on BM25 \cite{robertson2009probabilistic}, pretrained embeddings of words and entities obtained from \cite{yamada2020wikipedia2vec} (denoted as WikiVec) and BLINK \cite{wu-etal-2020-scalable}. (\RNum{3}) \textbf{Baseline of visual modality:} we retrieve the TopK candidate entities with the most similar visual contexts of the mention based on \textbf{CLIP} \cite{radford2021learning}.

\subsection{Contrastive Entity Disambiguation}

\begin{figure}[t]
	\centering
	\includegraphics[width=0.9\columnwidth]{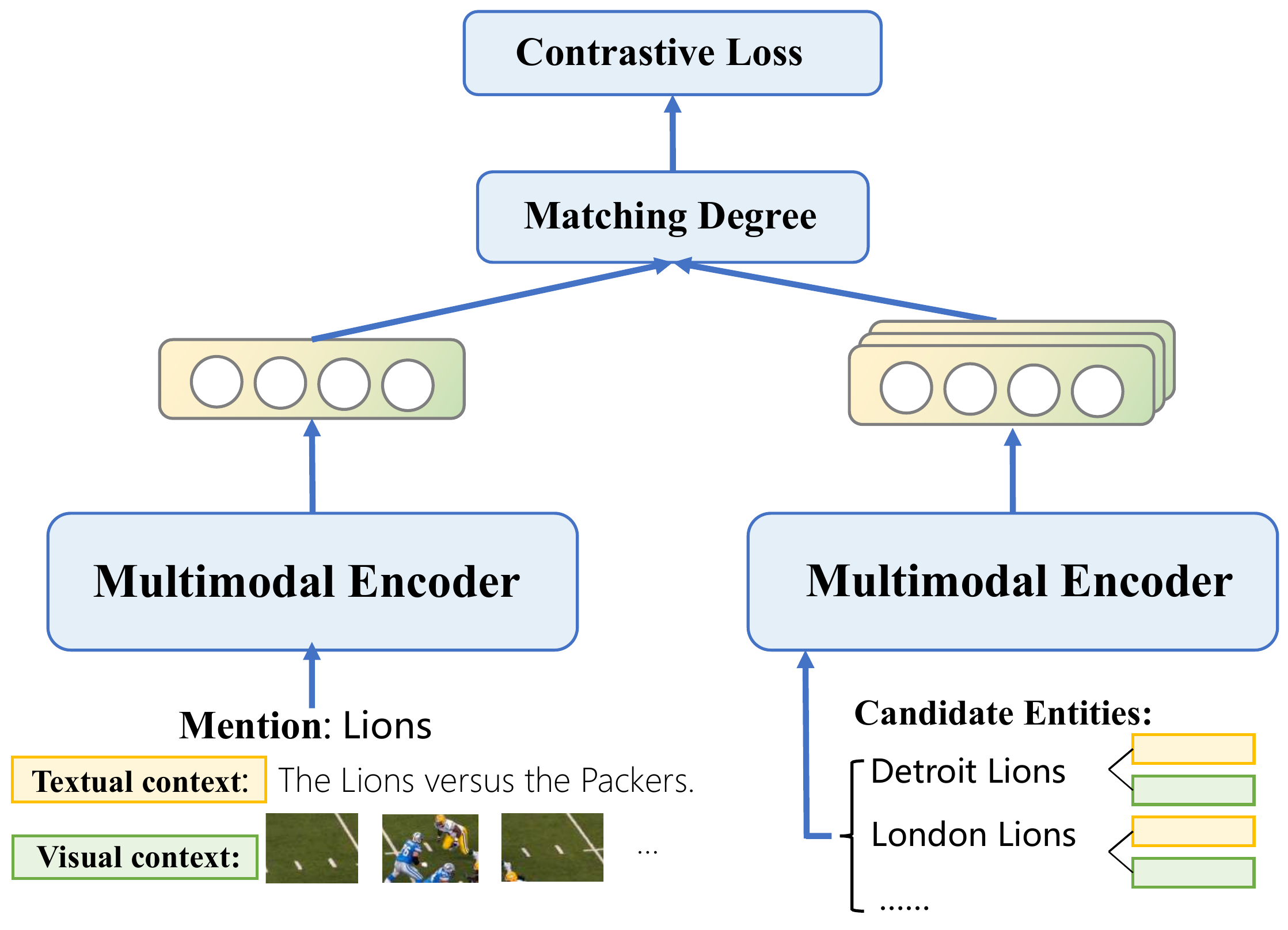}
	\caption{Framework of the introduced baselines.
}
	\label{fig:framework}
\end{figure}

The interaction between multimodal contexts of mentions and entities is complicated. It may bring noises to the model without careful handling. So we also introduce several baselines to explore the fusion of multimodal information.

The key component of ED is to design the function $\Psi(m;e_i)$ that quantifies the matching score between the mention $m$ and every entity $e_i\in E_m$. As shown in Figure~\ref{fig:framework}, the backbone of $\Psi(m;e_i)$ includes different multimodal encoders of $m$ and $e_i$ respectively, followed by dot-production to evaluate the matching degree between them. Specially, a multi-layer perceptron (MLP) is then used to combine the $P(e|m)$. Formally, $e^*$ of $m$ is predicted through:
\begin{equation}
\begin{aligned}
    \mathbf{m}=&\text{Encoder}_\text{m}(T_{m},V_{m});\mathbf{e_i} =\text{Encoder}_\text{e}(T_{e_i},V_{e_i})\\
    e^*&=\mathop{\arg\max}_{e_i\in E_m}\ \text{MLP}\left(\mathbf{m}\odot \mathbf{e_i}, P(e_i|m) \right)
\end{aligned}
\end{equation}
So the multimodal encoders of mentions and entities are the most significant parts of MEL. 
They use the same structure but training with different parameters.

\paragraph{Multimodal Encoder}

Firstly, we get the textual context's embeddings. For the mention's textual context $T_m=\{w_1,\dots,w_{L_1}\}$, we directly embed it with the word embedding layer of BERT \cite{devlin-etal-2019-bert}. While for $e_i$, we embed it as the pre-trained embeddings from \citet{yamada2020wikipedia2vec}, which have compressed the semantics of $e_i$'s entire contexts from Wikipedia.
\begin{equation}
    \{\hat{\mathbf{w}}_1,...,\hat{\mathbf{w}}_{L_1}\}=\text{BERT}_{EMB}(T_m)
\end{equation}

Secondly, we get the visual context embeddings. Instead of the widely used region-based visual features, we adopt grid features following \cite{huang2020pixel}, which has the advantage of end-to-end. 
Specifically, the visual features are represented with the grid features from :
\begin{equation}
    \{\hat{\mathbf{v}}_1,...,\hat{\mathbf{v}}_{L_2}\}=\text{Flat}(\text{ResNet}(V))
\end{equation}
where $\text{Flat}(\cdot)$ represents flatting the feature along the spatial dimension and $L_2$ indicates the number of grid features.

Finally, taking the embeddings of the two modalities as inputs, we capture the interaction between them. We adopt several backbones to fuse multiple modalities. 1) \textbf{UNITER} \cite{chen2020uniter}: the two modalities are concatenated and then fed into self-attention transformers to fuse them together. 2) \textbf{UNITER*}: we apply separate self-attention transformers to the two modalities before UNITER for better feature extraction of each modality. 3) \textbf{LXMERT} \cite{tan-bansal-2019-lxmert}: 
the two modalities are fed into separate self-attention transformers at first and then interact with cross-modal attention. The design of intra-modal and inter-modal attention helps better alignment and interaction of multiple modalities.

After multiple layers of the fusion operation: $\text{Fuse}\left(\{\hat{\mathbf{w}}_1,...,\hat{\mathbf{w}}_{L_1}\},\{\hat{\mathbf{v}}_1,...,\hat{\mathbf{v}}_{L_2}\}\right)$, the hidden states of the mention's tokens $\{\mathbf{h}_i,...,\mathbf{h}_j\}$ are obtained. Then we concatenate the hidden states of the first and the last tokens and feed them into a MLP to get the mention's embeddings: $\text{MLP}\left( [\mathbf{h}_i||\mathbf{h}_j]\right)$

\paragraph{Contrastive Loss} We introduce contrastive learning \cite{karpukhin-etal-2020-dense,gao2021simcse} to learn a more robust representation of both mentions and entities. It is widely acknowledged that selecting negative examples could be decisive for learning a good model. To this end, we utilize both hard negatives and in-batch negatives to improve our model's ability to distinguish between gold entities and hard/general negatives. Let $e_{i,j}$ represent the $j^{th}$ candidate entity of the $i^{th}$ mention in a batch and let $P_i$ denote the index of $m_i$'s gold entity. The hard negatives are the other $K-1$ candidate entities retrieved in CR step except for the gold entity: $\{e^-_{i,k}\}_{k\neq P_i}^{k\in[1, K]}$.  The in-batch negatives are gold entities of other $B-1$ mentions in the mini-batch: $\{e^+_{b, P_b}\}_{b\neq i}^{b\in[1, B]}$, where $B$ represents the batch size. The optimization objective is defined as the negative log likelihood of the ground-truth entity:
\begin{equation}
\begin{aligned}
    \mathcal{L}(m_i,&E_{m_i}) = -\log \frac{\mathrm{e}^{\Psi(m_i,e_{i,P_i}^+)}}{\mathrm{e}^{\Psi(m_i, e_{i,P_i}^+)}+\sum^{-}} \\
    \sum\nolimits^- = &\underbrace{\sum_{k=1,k\neq P_i}^{K}\mathrm{e}^{\Psi(m_i,e^-_{i,k})}}_{\text{hard negatives}}+\underbrace{\sum_{b=1,b\neq i}^{B}\mathrm{e}^{\Psi(m_i,e^+_{b, P_b})}}_{\text{in-batch negatives}}
\end{aligned}
\end{equation}

Besides the above baselines, we also compare with the following classic baselines: 1) \textbf {Baselines of Textual Modality} include {REL} \cite{le-titov-2018-improving}, {BERT} \cite{devlin-etal-2019-bert}, and {BLINK} \cite{wu-etal-2020-scalable}. 2) \textbf{Baselines of Visual Modality} include {ResNet-50} and {CLIP}. 3)  \textbf{Multimodal Baselines} include {MMEL18} \cite{moon-etal-2018-multimodal-named}, {MMEL20} \cite{10.1007/978-3-030-45439-5_31}. Details of the baselines can be found in the Appendix.

\section{Experimental Results}\label{sec:analysis}
\subsection{Candidate Retrieval Results}

\begin{table}[t]
\setlength{\tabcolsep}{0.65pt}
\centering
\begin{tabular}{clccc}
\toprule
\textbf{Modality} & \textbf{Method}  & \textbf{R@10} & \textbf{R@50} & \textbf{R@100} \\ \midrule
P & $P(e|m)$ & 80.82 & 85.48 & 86.23 \\
T & BM25 & 39.66 & 48.49 & 51.85 \\
T & WikiVec & 14.73 & 20.27 & 22.60 \\
T & BLINK & 63.63 & 73.15 & 76.03 \\
V & CLIP & 17.05 & 27.26 & 31.30 \\
T+V* & \scriptsize BLINK+CLIP & 66.96 & 77.18 & 80.53 \\
P+V* & \scriptsize $P(e|m)$+CLIP & 85.26 & 90.27 & 91.30 \\
P+T* & \scriptsize $P(e|m)$+BLINK & 86.36 & 91.78 & 93.21 \\
P+T+V* & \scriptsize $P(e|m)$+BLINK+CLIP & \textbf{86.37} & \textbf{91.91} & \textbf{93.35} \\ \bottomrule
\end{tabular}
\caption{Performance of candidate retrieval. R@K represents recall of the TopK retrieved entities. The modality of P, T, V represent the $P(e|m)$, textual context and visual context respectively. T+V and P+T+V represent the ensemble of different sub-methods. Results with * are generated using grid search over the Dev. dataset to find the best combination of different sub-methods.}
\label{tbl:candidate_retrieval}
\end{table}
\begin{figure*}[t]
\tiny
\centering
\includegraphics[width=0.75\textwidth]{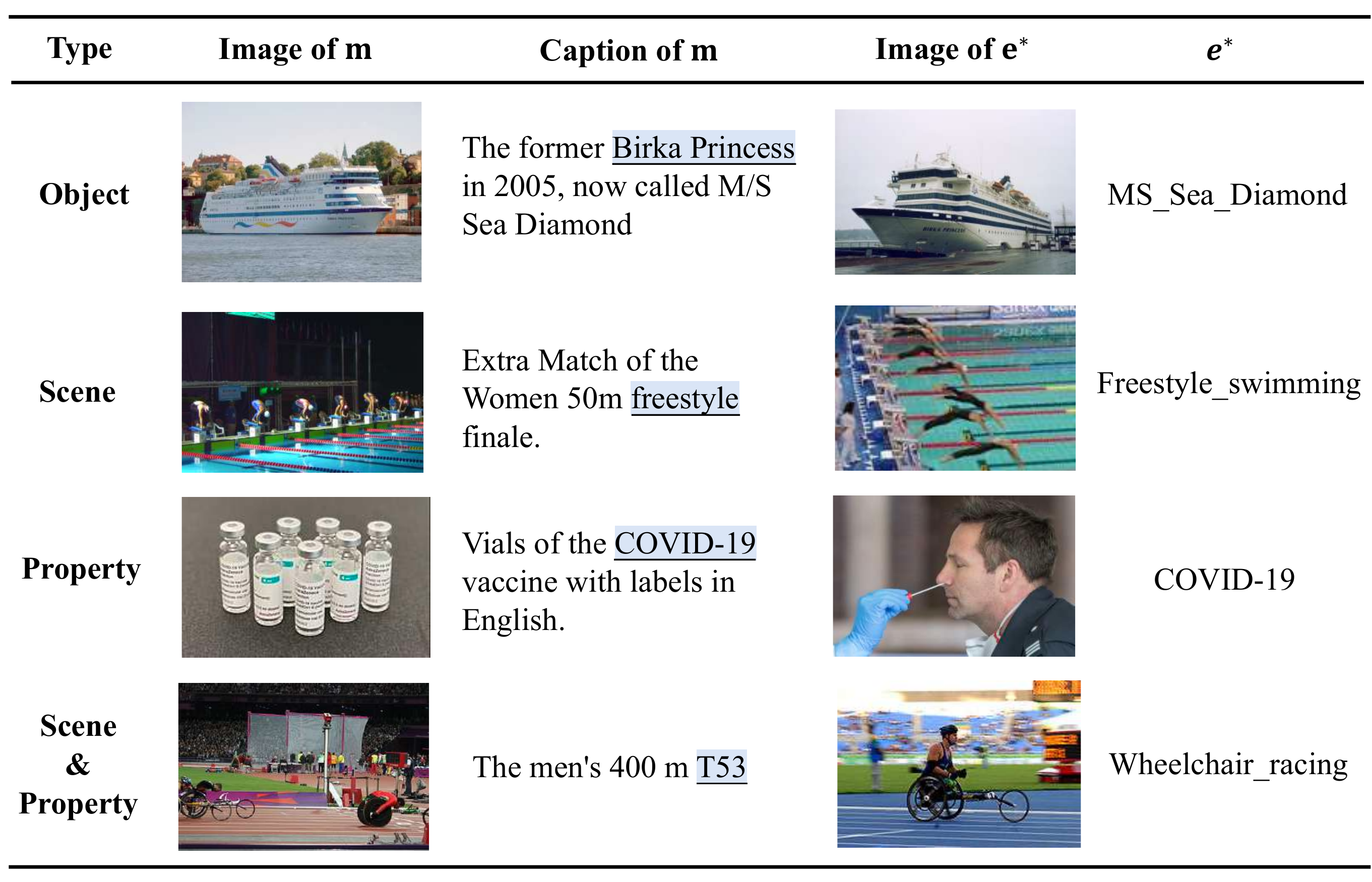}
\caption{Examples of the `Visual Clues'.}
\label{fig:visual_clues}
\end{figure*}
\begin{figure*}[t]
	\centering
	\includegraphics[width=0.8\textwidth]{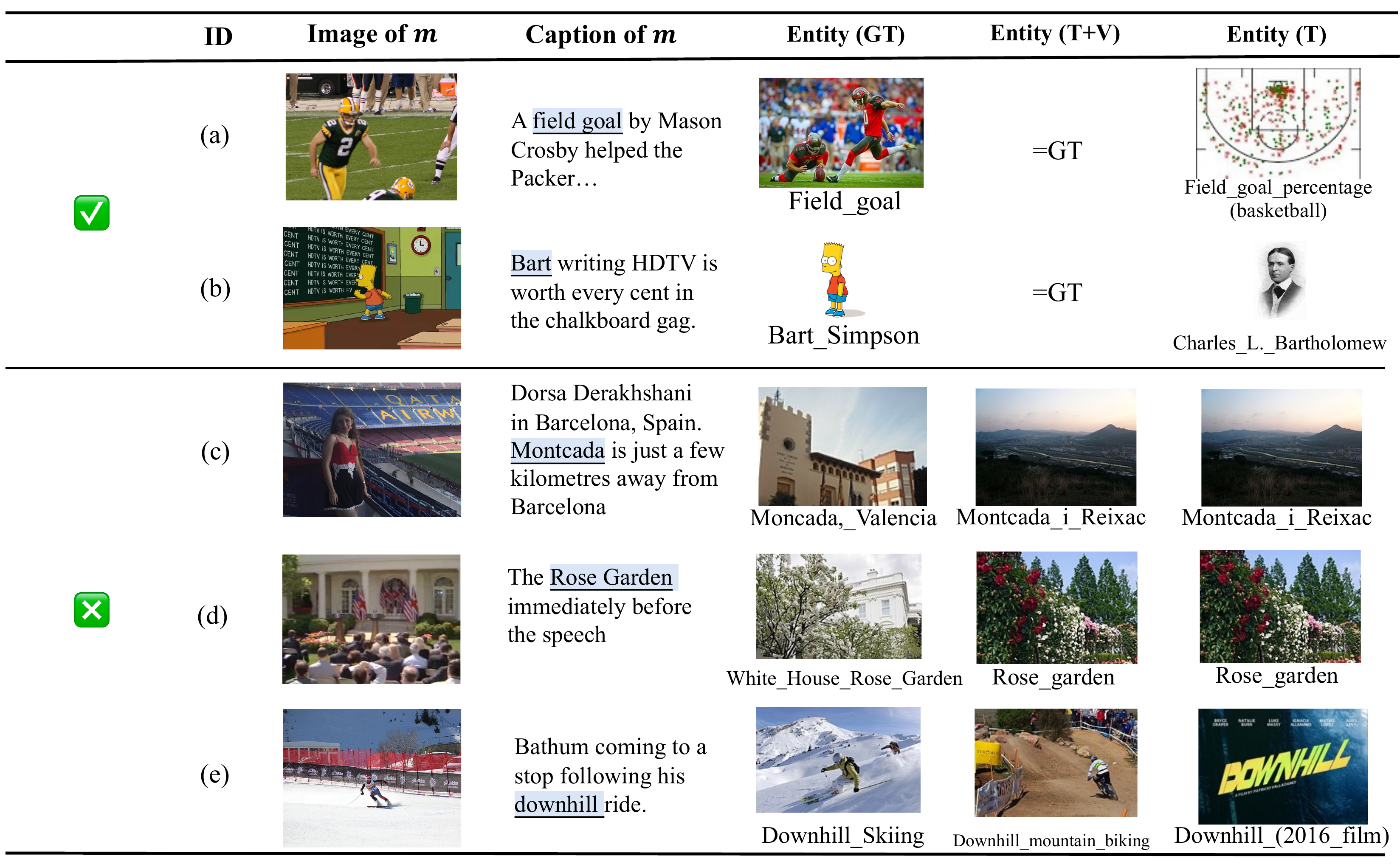}
	\caption{Case study. Successful predictions and failed predictions for the underlined mention are shown.}
	\label{fig:case_study}
	\vspace{2em}
\end{figure*}

As shown in Table ~\ref{tbl:candidate_retrieval}: 
1) Our model achieves 93.35\% of $R@100$, which indicates most related entities can be recalled from the large 16M KB. For retrieval, each mention takes about 12ms of P(e|m), 40ms of BM25, 183ms of WikiVec and CLIP, 60ms of BLINK;
2) As for ensemble of different modalities, T + V achieves better results than V and T, which verifies that the information of different modalities are complementary; 

In practice, we use grid search over the Dev. to find the best combination of different modalities. For example, when $K=10$, the best $E_m$ is generated with 80\%P+ 10\%T + 10\%V.

\subsection{Entity Disambiguation Results}
\begin{table}[t]
\setlength{\tabcolsep}{3pt}
\centering
\begin{tabular}{clccc}
\toprule
\bf Modality & \bf Model & \bf F1 & \bf P & \bf R \\ \midrule
\multirow{3}{*}{ T$\rightarrow$T } &  REL & 60.48 & 65.37 & 56.33 \\
 & BLINK & 66.74 & 70.93 & 63.03 \\
 & BERT & 63.65 & 69.63 & 58.77 \\
\hline
\multirow{2}{*}{ V$\rightarrow$V} & ResNet-50 & 40.16 & 43.81 & 37.08 \\
 & CLIP & 45.46 & 50.51& 41.33 \\ \hline
 T+V$\rightarrow$ T & MMEL18 & 61.58 & 70.85 & 54.46 \\ \hline
\multirow{7}{*}{ T+V$\rightarrow$ T+V} &  MMEL20 & 37.44 & 38.48 & 36.46 \\
 & UNITER & 69.37 & 73.72 & 65.51 \\
 & UNITER* & 70.60 & 75.03 & 66.66\\
 & LXMERT & 68.56 & 74.78 & 63.30\\ \cline{2-5} 
 & UNITER \dag & 71.07 & 75.52 & 67.10 \\
 & UNITER* \dag & 71.15 & 75.61 & 67.18 \\
 & LXMERT \dag & \bf 71.07 & \bf 78.62 & \bf 66.55 \\ \bottomrule
\end{tabular}
\caption{Comparison with baselines with results averaged over 5 runs. Models with \dag\  are enhanced with contrastive learning. All the models use the same candidate entity set retrieved through $P(e|m)$+BLINK+CLIP with $K=10$.}
\label{tbl:main_epx}
\end{table}

Following previous work, we report micro $F_1$, precision, recall in Table~\ref{tbl:main_epx}. According to the experimental results, we can see that: First, the proposed multimodal methods outperform all the methods with a single modality, which benefit from multimodal contexts. Besides, contrastive learning can even improve the performance. We reckon that contrastive learning improves the ability to distinguish entities.
Second, the textual baselines perform better that the visual ones, which indicates the textual context still plays a dominant role in MEL. 
Third, the methods using transformers to model the interaction between modalities perform better than those with simple interaction \cite{moon-etal-2018-multimodal-named,adjali-etal-2020-building}, which verifies the importance of fusing different modalities.

\subsection{Multimodal Analysis}
 
We also conduct some experiments on the ED tasks as following.

\begin{figure}[t]
	\centering
	\includegraphics[width=0.25\textwidth]{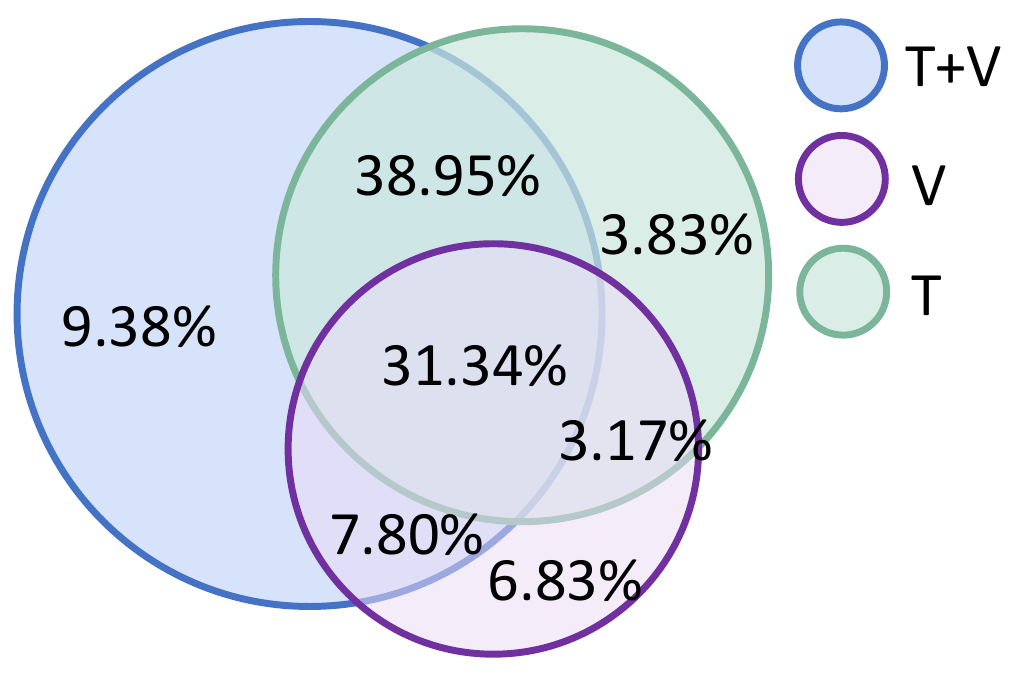}
	\caption{Venn diagram illustration of contributions of different modalities. We remove the input of the corresponding modality of LXMERT to get the results without re-training the model. To avoid the interference of $P(e|m)$, we also remove it from the model.
}
	\label{fig:venn_diagram}
\end{figure}

\paragraph{Are the multiple modalities complementary?} We draw a Venn diagram of different modalities in Figure~\ref{fig:venn_diagram}. The circle of Method $i$ is calculated through $\frac{\# Hit_{i}}{|Dataset|}$ and the interaction of two circles are calculated through $\frac{\# \left(Hit_{i} \cap Hit_{j} \right)}{|Dataset|}$. One can see that the textual modality is dominant, while the visual modality provides complementary information. Specially, the multimodal method predicts more new entities of 9.38\%, which verifies the importance of fusing two modalities.

\paragraph{Is it better to have multimodal contexts of both mentions and entities?} We conduct an ablation study and report the experimental results in Table~\ref{tbl:alba}. We can see that the model with multimodal contexts of both mentions and entities achieves the best result. So linking multimodal mentions to multimodal entities is better than linking multimodal mentions to mono-modal entities as done in \cite{moon-etal-2018-multimodal-named}.
\begin{table}[t]
\centering
\setlength{\tabcolsep}{6pt}
\small
\begin{tabular}{lccc}
\toprule
\bf Model & \bf F1 & \bf P & \bf R \\ 
\midrule
LXMERT & 71.07 & 78.62 & 66.55 \\ \hline
w/o $V_m$ & 53.75 & 58.62 & 49.62  \\
w/o $V_e$ & 67.04 & 73.12 & 61.89\\
w/o $V_m$ and $V_e$ & 63.65 & 69.63 & 58.77 \\ \hline
w/o $T_m$ & 59.76 & 65.19 & 55.17 \\
w/o $T_e$ & 51.30 & 55.95 & 47.36 \\
w/o $T_m$ and $T_e$ & 40.16 & 43.81 & 37.08 \\
\bottomrule
\end{tabular}
\caption{Ablation study to analyze modality absence of mention and entity. W/o $T_{m/e}$ or $V_{m/e}$ stands for LXMERT trained without the corresponding inputs.} 
\label{tbl:alba}
\end{table}

\paragraph{What visual clues are provided by the visual contexts?} We randomly select 800 image-caption pairs from the test dataset, and then ask annotators to label each mention with the types of visual clues. The visual clues include 4 types: 1) \textbf{Object}: the image contains the entity object. 2) \textbf{Scene}: the image reveals the scene that the entity belongs to (e.g. a basketball player of the `basketball game' scene). 3)  \textbf{Property}: the image contains some properties of the entity (e.g. an American flag reveals the property of a person's nationality). 4)  \textbf{Others}: other important contexts. Note that the four types of clues can be crossed and a sample could have no clues. Examples of the visual clues can be found in Figure~\ref{fig:visual_clues}. We find that visual context is helpful for 60.54\% mentions and 81.56\% image-caption pairs. We report the contribution of different types of visual clues in Table~\ref{tbl:image_type}. One can see that: 1) For scene clues, object clues and property clues, the T+V significantly outperforms T. It demonstrates that the multimodal model benefits a lot from the information of multiple types of visual clues in the images. 2) But our model still does not perform well with the scene and property clues. So fine-grained visual clues are not used well and this indicates the direction of future research.

\begin{table}[t]
\centering
\small
\begin{tabular}{lccc}
\toprule
\multirow{2}{*}{\textbf{Visual Clues}} & \multirow{2}{*}{\textbf{Proportion}} & \multicolumn{2}{c}{$\mathbf{F_1}$} \\ 
\cline{3-4} 
 &  & \textbf{T}  & \textbf{T+V}  \\ 
\midrule
Object & 45.40\% & 59.65 & 67.72  \\
Scene & 18.96\% & 49.68 & 60.63  \\
Property & 26.22\% & 56.28 & 64.45  \\
Others & 14.80\% & 58.82 & 88.24  \\ 
\bottomrule
\end{tabular}
\caption{Model performance under different visual clues. T+V denotes the multimodal model LXMERT, and T represents the textual model BERT.} 
\label{tbl:image_type}
\end{table}

\subsection{Case Study}
We present several examples where multimodal contexts influence MEL in Figure~\ref{fig:case_study}. Example~(a) and (b) verify the helpfulness of the multimodal context. From the error cases, we can see that the model still lacks such capabilities: 1) Eliminate the influence of unhelpful images (e.g., Example~(c)); 2) Perform reasoning (e.g., inferring the ``white house'' from Example~(d)'s image); 3) Alleviate over-reliance on $P(e|m)$ (e.g., Example~(e)).

\section{Conclusion and Future Work}\label{sec:conclusion}
We propose \textsc{Wiki}Diverse, a manually-annotated Wikipedia-based MEL dataset collected from Wikinews. To overcome the weaknesses of existing datasets, \textsc{Wiki}Diverse covers a wide range of topics, entity types and ambiguity. We implement a series of baselines and carry out multiple experiments over the dataset. 
According to the experimental results, \textsc{Wiki}Diverse is a challenging dataset worth further exploration. 
Besides multimodal entity linking, \textsc{Wiki}Diverse can also be applied to evaluate the pre-trained language model, multimodal named entity typing/recognition, multimodal topic classification, etc.
In the future, we plan to 1) utilize more than one images of each entity 2) adopt finer-grained multimodal interaction models for this task and 3) transfer the model to more general scenarios such as EL in articles.

\section*{Acknowledgement}
This research was supported by the National Key Research and Development Project
(No. 2020AAA0109302), National Natural Science Foundation of China (No. 62072323),  Shanghai Science and Technology Innovation
Action Plan (No. 19511120400), Shanghai Municipal Science
and Technology Major Project (No. 2021SHZDZX0103) and Alibaba Research Intern Program. 

\bibliographystyle{acl_natbib}

\end{document}